# Theorem Proving Based on Semantics of DNA Strand Graph

KUMAR S. RAY
Electronics and Communication Science Unit
Indian Statistical Institute
203, B.T Road, Kolkata-700108, India
E-mail: ksray@isical.ac.in
Tel: +91 8981074174

MANDRITA MONDAL
Electronics and Communication Science Unit
Indian Statistical Institute
203, B.T Road, Kolkata-700108, India
E-mail: mandritamondal@gmail.com
Tel: +91 9830354798

**Abstract**
Because of several technological limitations of traditional silicon based computing, for past few years a paradigm shift, from silicon to carbon, is occurring in computational world. DNA computing has been considered to be quite promising in solving computational and reasoning problems by using DNA strands. Resolution, an important aspect of automated theorem proving and mathematical logic, is a rule of inference which leads to proof by contradiction technique for sentences in propositional logic and first-order logic. This can also be called refutation theorem-proving. In this paper we have shown how the theorem proving with resolution refutation by DNA computation can be represented by the semantics of process calculus and strand graph.

**Keywords:** theorem proving, resolution refutation, strand graph, process calculus, DNA computing, propositional logic, proof by contradiction, strand displacement.

**1. Introduction**

Though traditional silicon based computing has widely been used for past several decades, it has a number of technological challenges in terms of requirement of memory, energy consumption, density and heat dissipation. In spite of the flexibility of this conventional technology, in some aspect, it has reached its limitations of design complexity and processing power. For miniaturization of the disadvantages, in modern age several alternatives to traditional silicon based technology have been proposed. In recent past DNA computing has been considered to be quite promising in solving computational and reasoning problems by using DNA strands which is a powerful tool for engineering at nano-scale [Adleman, 1994; Winfree et al., 1998; Benenson et al., 2001; Chang and Gou, 2003; Green et al., 2006; Akerkar and Sajja, 2009]. Thus, it can be said that a paradigm shift, from silicon to carbon, is occurring in computational world. The behaviour of DNA strands can be manipulated by standard DNA operations, and by setting up the strands of DNA in the right way, logical reasoning and decision making can be predicted [Yeung and Tsang, 1997; Ray and Mondal, 2011a; Ray and Mondal, 2011b; Ray and Mondal, 2016].

In this paper we have shown how the theorem proving with resolution refutation by DNA computation can be presented by the semantics of DNA strand graph.

Resolution, an important aspect of automated theorem proving and mathematical logic, can be defined as a rule of inference which leads to proof by contradiction technique for sentences in propositional logic and first-order logic. Proof by contradiction can also be called refutation theorem-proving. When two clauses contains complementary literals, a valid rule of resolution generates a new clause from these two clauses. A propositional variable or its negation (i.e., P, ¬P) is called a literal. Resolution is the only interference rule which needs to build a complete theorem prover, based on proof by contradiction and usually called resolution refutation [Chang and Lee, 1997].

In this paper we will use formal language theory as a tool of modeling and analysis of DNA operations performed for theorem proving in propositional logic. Though computing language is a complex task, formal language theory has taken the advantage of the idea of defining semantics and formalizing architecture of the wet lab procedure. In their research work Petersen, Lakin and Phillips [Petersen *et. al.*, 2016] developed a domain-specific DNA strand displacement (DSD) language for modeling, simulating and analyzing DNA strand displacement systems. Different types of DNA structures are used for computation and reasoning. Thus, more general formal language is required which can encode arbitrary secondary structures of DNA strands and their interactions. Petersen *et. al.* [Petersen *et. al.*, 2016] again redefine the syntax and semantic of the DSD language to extend the scope of the language. The proposed reformulated language is termed as *process calculus*. The expressive syntax and formal semantics of process calculus can model, simulate and analyze the mechanism of strand displacement of DNA strands with rich secondary structures such as branches and loops. The complex formal models used to solve reasoning and computation problems can have corresponding graphical representation which is defined as *strand graph* [Petersen *et. al.*, 2016].

## 2. The Resolution Principle in Propositional Logic

*Theorem proving*, a subfield is of automated reasoning and mathematical logic, is used to develop computer programs. It shows that some statement, i.e. conjecture, is a logical consequence of a set of hypotheses. Theorem proving is applicable for several domains. In this paper we will perform theorem proving with resolution refutation in propositional logic [Chang and Lee, 1997].

A *proposition* is an assertion which is either true or false but not both. *Propositional variable* denotes arbitrary propositions with unspecified truth value such as *P*, *Q*, *R*. These variables can be connected with *logical connectives*, for example, *and* (*conjunction* ∧), *or* (*disjunction* ∨), *not* (*negation* ¬). A propositional variable or its negation is called a *literal*. For example, if *P* is a propositional variable, then *P* and ¬*P* are both literals. An assertion which contains at least one propositional variable is called to be in *propositional form*. *Propositional logic*, the branch of logic, is the study of propositions that are formed by other propositions by logical connectives. Propositional logic is also concerned on how their value depends on the truth value of their components. Apart from the above mentioned logical operators there are two more operators which are used in logic. One is called *implication* ($\Rightarrow$) and other is *equivalence* ($\Leftrightarrow$).

*Propositional resolution* which is a rule of inference, is capable to generate theorem prover in the domain of propositional logic. Before the application of resolution principle in propositional logic, the premises and conclusions must be expressed in *clausal form*. A *clausal sentence* is either a literal or a disjunction of literals. If $P$ and $Q$ are propositional variable, then the clausal sentences are:

$$P$$
$$\neg P$$
$$\neg P \vee Q$$

A *clause* is the set of literals in a clausal sentence. The clauses of above mentioned clausal sentences are:

$$\{P\}$$
$$\{\neg Q\}$$
$$\{\neg P, Q\}$$

The *empty set* { } is also a clause. It is equivalent to an empty disjunction and, therefore, is unsatisfiable. Thus, the clausal form and clauses in propositional logic can be defined as follows:

$$(clausal\ form) := (clause) \wedge (clause) \wedge \cdots \wedge (clause)$$
$$(clause) := (literal) \vee (literal) \vee \cdots \vee (literal)$$

The rules for conversion of arbitrary set of propositional logic sentences to equivalent set of clauses are given below:

1. Implications:

    $P \Rightarrow Q$ $\quad\rightarrow\quad$ $\neg P \vee Q$
    $P \Leftarrow Q$ $\quad\rightarrow\quad$ $P \vee \neg Q$
    $P \Leftrightarrow Q$ $\quad\rightarrow\quad$ $(\neg P \vee Q) \wedge (P \vee \neg Q)$

2. Negations:

    $\neg\neg P$ $\quad\rightarrow\quad$ $P$
    $\neg(P \wedge Q)$ $\quad\rightarrow\quad$ $\neg P \vee \neg Q$
    $\neg(P \vee Q)$ $\quad\rightarrow\quad$ $\neg P \wedge \neg Q$

3. Distribution:

    $P \vee (Q \wedge R)$ $\quad\rightarrow\quad$ $(P \vee Q) \wedge (P \vee R)$
    $(P \wedge Q) \vee R$ $\quad\rightarrow\quad$ $(P \vee R) \wedge (Q \vee R)$
    $P \vee (P_1 \vee ... \vee P_n)$ $\quad\rightarrow\quad$ $P \vee P_1 \vee ... \vee P_n$
    $(P_1 \vee ... \vee P_n) \vee P$ $\quad\rightarrow\quad$ $P_1 \vee ... \vee P_n \vee P$

$$P \wedge (P_1 \wedge ... \wedge P_n) \rightarrow P \wedge P_1 \wedge ... \wedge P_n$$
$$(P_1 \wedge ... \wedge P_n) \wedge P \rightarrow P_1 \wedge ... \wedge P_n \wedge P$$

4. Operators (O):

$$P_1 \vee ... \vee P_n \rightarrow \{P_1, ... , P_n\}$$
$$P_1 \wedge ... \wedge P_n \rightarrow \{P_1\}, ... , \{P_n\}$$

*Resolution principle* states that: *"For any two clauses $C_1$ and $C_2$, if there is a literal $L_1$ in $C_1$ that is complementary to a literal $L_2$ in $C_2$, then delete $L_1$ and $L_2$ from $C_1$ and $C_2$, respectively, and construct the disjunction of the remaining clauses. The constructed clause is a resolvent of $C_1$ and $C_2$."* [Chang and Lee, 1997]

For example, let,

$C_1$:    $P$
$C_2$:    $\neg P \vee Q$.

According to the resolution principle, the complementary pair of literal, i.e. $P$ in $C_1$ and $\neg P$ in $C_2$, should be deleted to construct the resolvent $C_3$. The resolvent $C_3$ is:

$C_3$:    $Q$.

Another example is given below,

$C_1$:    $\neg Q \vee R$
$C_2$:    $\neg P \vee Q \vee \neg S$

The resolvent of $C_1$ and $C_2$ is:

$C_3$:    $R \vee \neg P \vee \neg S$

If there is no complementary literal in $C_1$ and $C_2$, the no resolvent can be constructed from given clauses. For example,

$C_1$:    $\neg R \vee S$
$C_2$:    $\neg R \vee Q \vee T$

Another property of resolution principle is, *"if two clauses $C_1$ and $C_2$ are given, a resolvent C of $C_1$ and $C_2$ is a logical consequence of $C_1$ and $C_2$".*[ Chang and Lee, 1997]

We have previously mentioned that, if the resolution principle generate empty clause { } from a set of clauses *S*, then it can be said that *S* is unsatisfiable. The following definition can be drawn from the principle of resolution:

*"Given a set of clauses S, a (resolution) deduction of C from S is a finite sequence $C_1$, $C_2$, ..., $C_k$ of clauses such that each $C_i$, either is a clause in S or a resolvent of clauses preceding $C_i$, and $C_k$ = C. A deduction of {} from S is called a refutation, or a proof of S."* [Chang and Lee, 1997]

Thus, the resolution principle can be used to prove the unsatisfiability of a set of clauses. This can be explained by the following examples.

Let *S* is a set containing six clauses,

$(i)\ P \lor \neg Q \lor R$
$(ii)\ \neg U \lor V \lor \neg R$
$(iii)\ Q$
$(iv)\ \neg V$
$(v)\ \neg P$
$(vi)\ U$

$\left.\right\}\ S$

From (i) and (iii), the generated resolvent is,
$(vii)\ P \lor R$

From (ii) and (iv), the generated resolvent is,
$(viii)\ \neg U \lor \neg R$

From (vii) and (viii), the generated resolvent is,
$(ix)\ P \lor \neg U$

From (ix) and (v), the generated resolvent is,
$(x)\ \neg U$

From (x) and (vi), the generated resolvent is,
$(xi)\ \{\}$

Since {} is derived from the set of clauses $S$ by resolution, it can be said that the empty clause {} is the logical consequence of $S$. {} can only be a logical consequence of an unsatisfiable set of clauses. Hence, it is proved that $S$ is unsatisfiable. Fig. 1 shows the *deduction tree* of the above deduction.

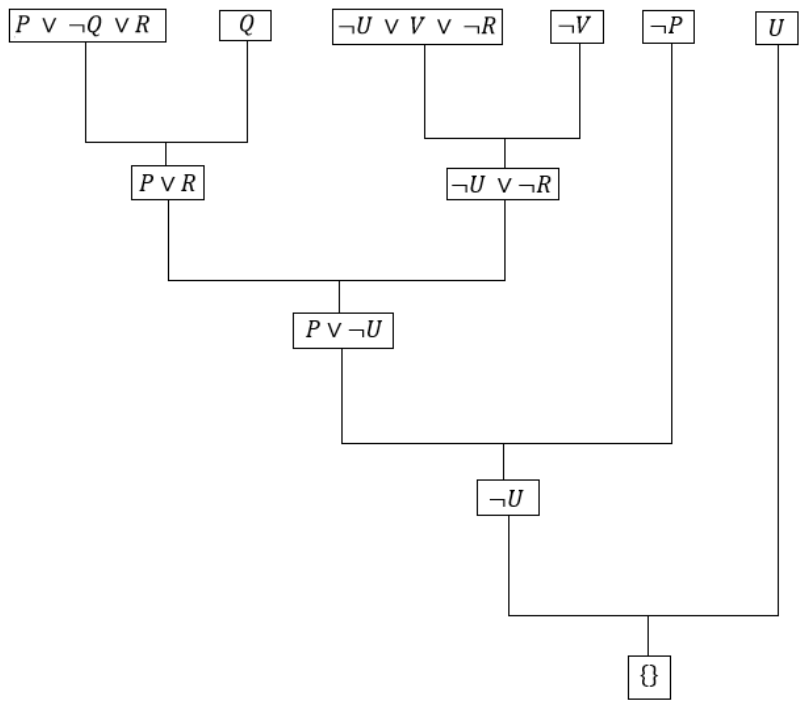

Figure 1. Deduction tree

Thus, it can be said that resolution refutation or *proof by contradiction* proves a theorem by negating the statement to be proved. The negated statement is added to the set of premises which are known to be true. The theorem prover, generated by propositional resolution, proves the consistency of the negated goal. The inconsistency of the negated goal with the given set of premises implies that the original goal is consistent.

Let, we want to prove a premise or axiom *X* from a set of axioms *Z*. The general steps of resolution refutation for proving *X* is given below:

*Step 1.* All the premises or axioms of *Z* are expressed in clausal form. The set of clauses are denoted by *S*.

*Step 2.* In the set of axioms expressed in clausal form negation of what is to be proved is added. Here, negation of *X* is added.

*Step 3.* All these clauses are resolved together, producing new clauses, i.e. resolvents, which logically follow from them.

*Step 4.* If the resolvent is an empty clause, a contradiction is generated. The process should be stopped as *X* is proved to be true.

*Step 5.* Else if no resolvent can be generated, again stop the process; as *X* is proved to be false.

*Step 6.* Else the resolvent is added to *S* and step 3 is repeated.

In this paper the theorem proving with resolution refutation by DNA strands are represented by the formal language of process calculus and strand graph semantics. The next section shows the syntax and semantics of process calculus and strand graph. [Petersen *et. al.*, 2016]

## 3. Syntax and Semantics of Process Calculus and Strand Graph

In DNA computing DNA strands are used to perform computation. DNA strands are the strings containing four DNA bases i.e. A, T, G, C. Formal language theory, which deals with the DNA strands, is used to model, simulate and analyze the concurrent communicating processes of DNA computation. There is a resemblance between generative grammar of formal language theory and the self-assembly and ligation of the DNA strands. Both generate new strings from previous string following some pre-defined rule. Thus, to represent the architecture of a model of DNA computation formal language is widely appreciated. The mechanism of strand displacements in DNA strands with rich secondary structures can be modeled, simulated and analyzed by a newly defined language by Petersen, Lakin and Phillips [Petersen *et. al.*, 2016], termed as *process calculus*. Now we will discuss the formal syntax and semantics of process calculus to formulate the architectures of DNA models.

A language is a set of valid sentences. The validity of language can be broken down into two things: syntax and semantics. The term syntax refers to grammatical structure of a language and the term semantics is concerned to the meaning of the vocabulary symbols arranged with that structure, often in relation to their truth and falsehood. Grammatical (syntactically) valid does not imply sensible (semantically) valid. In mathematics, computer science and linguistics, a formal language is a set of strings of symbols that may be constrained by rules that are specific to it [Ray and Mondal, 2016].

In process calculus, a process or program $P$ is defined as a multiset of DNA strands $<S>$.

Process or program $P ::= <S_1> |...| <S_i>$      where, $i \geq 0$

Each strand $<S>$ contains one or more domains $d$. Domain is actually a sequence of DNA bases or nucleotides i.e. A, T, G, C.

Strand $S ::= d_1 ..... d_i$      where, $i \geq 0$

A domain $d$ in a DNA strand is either free or bound with the complementary domain of any other DNA strand or to the same strand. A free domain is denoted by $d$. If the domain is bound by bond $x$ them the bound domain is denoted by $d!x$. Let, an arbitrary domain is named $r$, then $r*$ is the complementary domain to which $r$ can bind by Watson-Crick base pairing. A domain is called *toehold* $t\textasciicircum$ if it is short enough to spontaneously unbind from its complement $t\textasciicircum*$.

The semantics of process calculus depends on some functions which determine whether a rule can be applied on a program. The functions are listed below;

- The function *comp(r)* returns the complementary domain of domain *r*. Thus, it can be said that, *comp(r) = r\** and *comp(r\*) = r*.
- The function *toehold(r)* returns true if *r* is a toehold domain. Then we can also represent domain *r* by *r^*.
- The function *adjacent(x, P)* returns the set of bonds that are adjacent to bond *x* in program *P*.
- The function *hidden(x, P)* returns true if one end of bond *x* occurs within a closed loop. Thus, the specific domain cannot bind to its complementary sequence.
- The function *anchored(x, P)* returns true if both ends of bond *x* are held "close" to each other. Thus, bond *x* is a part of a stable junction.
- The *context* $C(S_1, ..., S_i)$ is defined as a process *P* containing sequences $S_1, ..., S_i$.
- The function *permute($S_1, ..., S_i$)* returns any possible permutation of sequences $S_1, ..., S_i$.

Now, we will define the semantics of some rules of process calculus by the following figures and corresponding expression.

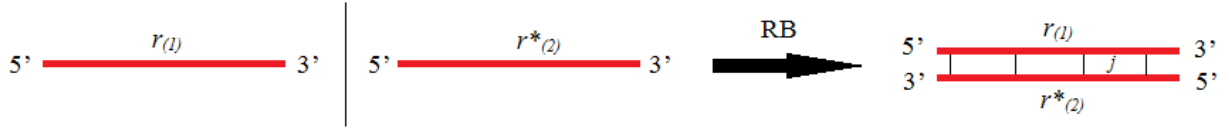

Figure 2. Rule (RB)

The semantics of rule (RB) as shown in Fig. 2 can be presented as,

$$(RB) \quad \frac{\neg hidden(j, P)}{C(r, r^*) \xrightarrow{RB, \{j\}} C(r!j, r^*!j) = P}$$

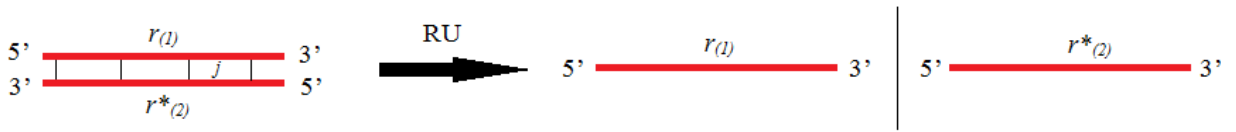

Figure 3. Rule (RU)

The semantics of rule (RU) as shown in Fig. 3 can be presented as,

$$(RU) \quad \frac{\neg anchored(j, P) \quad toehold(r)}{P = C(r!j, r^*!j) \xrightarrow{RU, \{i\}} C(r, r^*)}$$

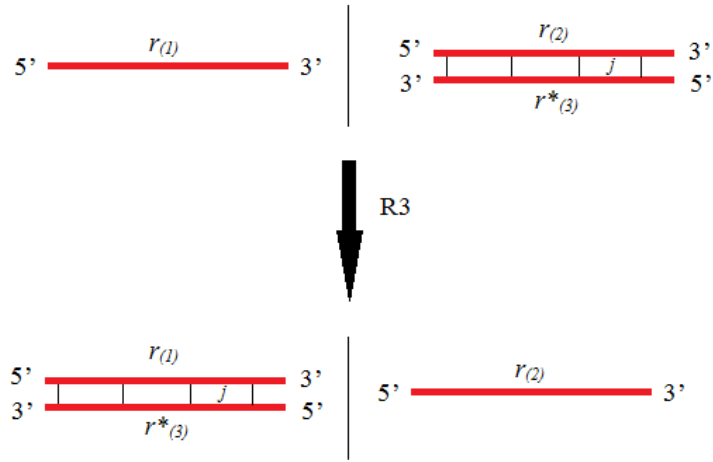

Figure 4. Rule (R3)

The semantics of rule (R3) as shown in Fig. 4 can be presented as,

$$(RU) \quad \frac{anchored(j, P)}{C(r, r!j, r^*!j) \xrightarrow{R3,\{j\}} C(r!j, r, r^*!j) = P}$$

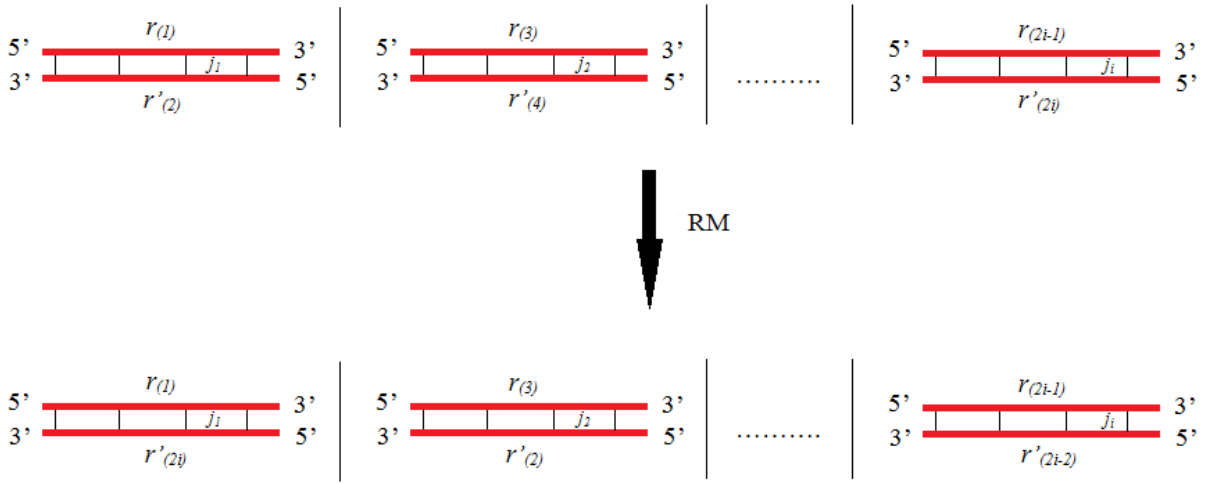

Figure 5. Rule (RM)

The semantics of rule (RM) as shown in Fig. 5 can be presented as,

$$(RM) \quad \frac{anchored(j_1, P') \ldots \ldots anchored(j_i, P')}{C(r!j_1, r'!j_1, r!j_2, r'!j_2, \ldots .. r!j_i, r'!j_i) \xrightarrow{RM,\{j_1 \ldots j_i\}} C(r!j_1, r'!j_2, r!j_2, r'!j_3, \ldots .. r!j_i, r'!j_1) = P'}$$

Now we will illustrate the reduction rules of process calculus by the help of an example. We take hairpin toehold exchange program with two invader strands as the example. The pictorial representation of the program is shown in Fig. 6.

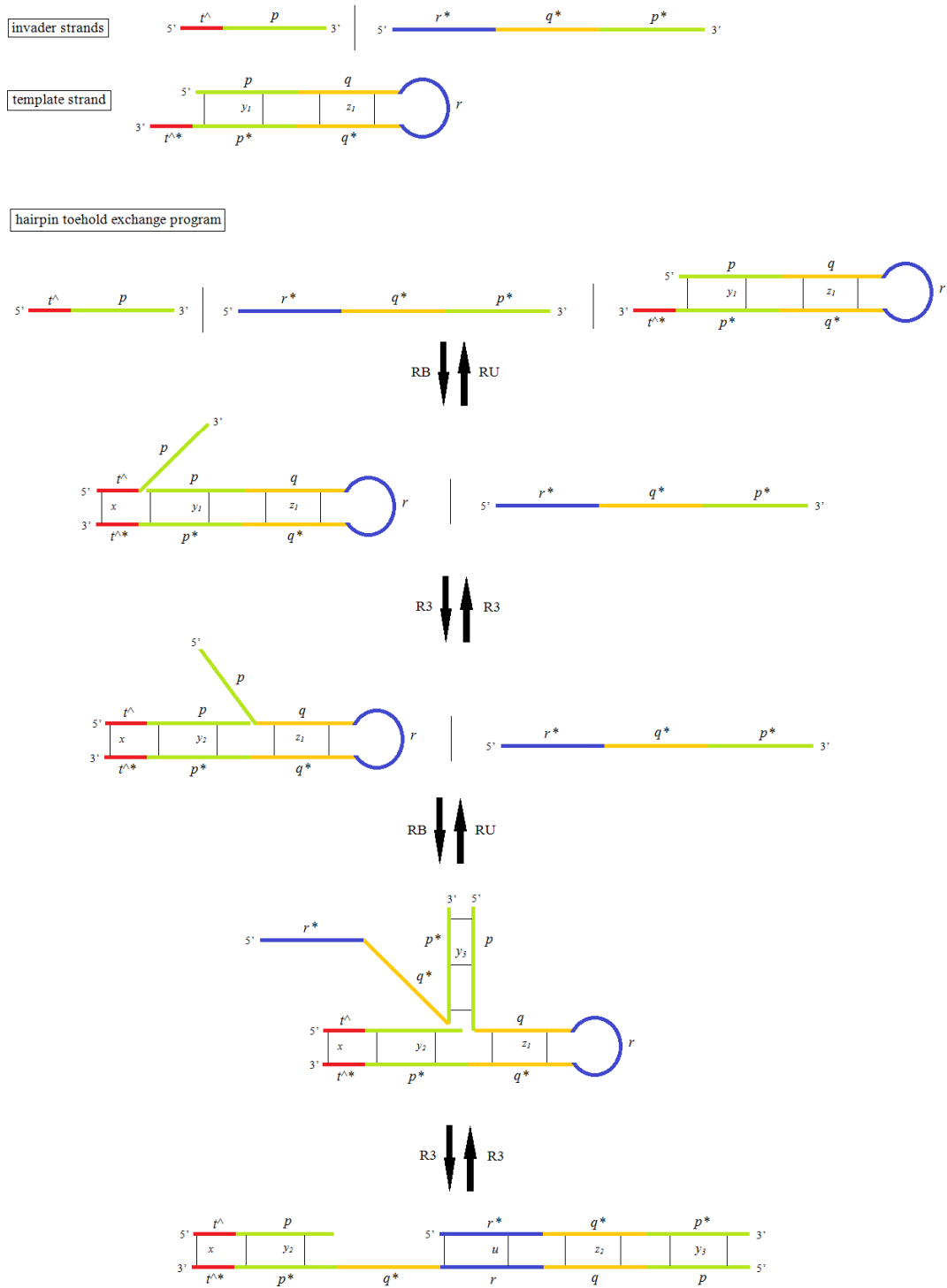

Figure 6. Hairpin toehold exchange program with to invader strands

In the illustrated example there are two invader strands and one template strands. One of the two single stranded invader strands has two domains ($t\hat{}$, $p$) and the other has three domains ($r^*$, $q^*$, $p^*$). The template strand with secondary hairpin structure contains five domains ($p$, $q$, $r$, $q^*$, $p^*$, $t^{\hat{}*}$). The program codes of the strands are given below;

$$invader = < t^{\hat{}}\ p > | \ invader = < r^*\ q^*\ p^* > | \ template = < p!y_1\ q!z_1\ r\ q^*!z_1\ p^*!y_1\ t^{\hat{}*} >$$

The function *toehold(t)* returns true for the single stranded invader strand. Thus, the domain at 5' end of the strand is denoted by $t\hat{}$. This domain has a free complementary domain $t^{\hat{}*}$ in the template strand as the program matches the context $C(t, t^*)$. It can be written that $P=C(t^{\hat{}}!x, t^{\hat{}*}!x)$ as one end of the bond $x$ is not in closed loop, i.e. *hidden(x, P)* returns false. Thus, the program $P'$ can be produced by the *rule (RB)* which forms the new bond $x$ between the single stranded invader strand and the template. The program code is shown below;

$$< t^{\hat{}}\ p > | < r^*\ q^*\ p^* > | < p!y_1\ q!z_1\ r\ q^*!z_1\ p^*!y_1\ t^{\hat{}*} >$$
$$\xrightarrow{(RB)} < \boldsymbol{t^{\hat{}}!x}\ p > | < r^*\ q^*\ p^* > | < p!y_1\ q!z_1\ r\ q^*!z_1\ p^*!y_1\ \boldsymbol{t^{\hat{}*}!x} >$$

As domain $t$ is toehold, it is short enough to unbind spontaneously. Here the program *anchored(x, P)* returns false as the bond $x$ is not a part of a junction that holds both ends of the bond close to each other. Thus, the *rule (RU)* can also occur which breaks the bond $x$ between the single stranded invader strand and the template to produce the program $C(t, t^*)$. It is reversible of rule (RB). The program code is shown below;

$$< \boldsymbol{t^{\hat{}}!x}\ p > | < r^*\ q^*\ p^* > | < p!y_1\ q!z_1\ r\ q^*!z_1\ p^*!y_1\ \boldsymbol{t^{\hat{}*}!x} >$$
$$\xrightarrow{(RU)} < \boldsymbol{t^{\hat{}}}\ p > | < r^*\ q^*\ p^* > | < p!y_1\ q!z_1\ r\ q^*!z_1\ p^*!y_1\ \boldsymbol{t^{\hat{}*}} >$$

In the next step toehold mediated branch migration and strand displacement occurs. Strand displacement is the process through which two DNA strands with partial or full complementarity hybridize to each other, displacing one or more pre-hybridized strands [Zhang and Seelig, 2011]. The free domain $p$ of the invader strand has a complementary domain $p^*$ in the template strand which is already bound by the bond $y_1$. In this step the program matches the context $C(p, p!y_1, p^*!y_1)$. We have to check if an anchored bond can be formed between the invader strand and the template to produce the program $P' = C(p!y_2, p, p^*! y_2)$. The formation of the new bond $y_2$ is possible by applying *rule (R3)* as there is a bond $x$ that is immediately adjacent to $y_2$ in $P'$, holding both ends of bond $y_2$ close to each other.

$$< t^{\hat{}}!x\ \boldsymbol{p} > | < r^*\ q^*\ p^* > | < \boldsymbol{p!y_1}\ q!z_1\ r\ q^*!z_1\ \boldsymbol{p^*!y_1}\ t^{\hat{}*}!x >$$
$$\xrightarrow{(R3)} < t^{\hat{}}!x\ \boldsymbol{p!y_2} > | < r^*\ q^*\ p^* > | < \boldsymbol{p}\ q!z_1\ r\ q^*!z_1\ \boldsymbol{p^*!y_2}\ t^{\hat{}*}!x >$$

Next, the other invader strand comes in action. There is a free domain $p^*$ at 3' end of the invader strand. This domain has a free complementary domain $p$ in the 5' end of the template strand. The formation of a new bond $y_3$ is possible as one end of the bond does not occur in closed loop, i.e. the bond is not hidden. The formation of the bond according the *rule (RB)* is shown in the following program code;

$$< t\hat{}!x \quad p!y_2 > | < r^* \quad q^* \quad \boldsymbol{p^*} > | < \boldsymbol{p} \quad q!z_1 \quad r \quad q^*!z_1 \quad p^*!y_2 \quad t^{\hat{}*}!x >$$
$$\xrightarrow{(RB)} < t\hat{}!x \quad p!y_2 > | < r^* \quad q^* \quad \boldsymbol{p^*!y_3} > | < \boldsymbol{p!y_3} \quad q!z_1 \quad r \quad q^*!z_1 \quad p^*!y_2 \quad t^{\hat{}*}!x >$$

This reaction is a reversible reaction. The previous step can be restored by *rule (RU)*.

$$< t\hat{}!x \quad p!y_2 > | < r^* \quad q^* \quad \boldsymbol{p^*!y_3} > | < \boldsymbol{p!y_3} \quad q!z_1 \quad r \quad q^*!z_1 \quad p^*!y_2 \quad t^{\hat{}*}!x >$$
$$\xrightarrow{(RU)} < t\hat{}!x \quad p!y_2 > | < r^* \quad q^* \quad \boldsymbol{p^*} > | < \boldsymbol{p} \quad q!z_1 \quad r \quad q^*!z_1 \quad p^*!y_2 \quad t^{\hat{}*}!x >$$

The *p\** domain of the invader strand is not a toehold as it is a long domain and does not unbind spontaneously. But, when *p\** hybridize to the template strand; it leads to brand migration and strand displacement. This process breaks the unanchored bond $z_1$ and opens the hairpin of the template strand to produce the program *C(q, q\*)*. The anchored bonds, first $z_2$ then *u*, can be formed between the invader strand and the template. There is a bond $y_3$ immediately adjacent to $z_2$ in this process, holding the both ends of $z_2$ close to each other. Again after formation of $z_2$ between *q\** domain of the invader strand and its complementary domain in the template strand, it holds the both ends of bond *u* close to each other. Thus, new bond *u* is formed between *r\** domain of the invader strand and its complementary domain *r* in the template strand. This process occurs by applying *rule (R3)*.

$$< t\hat{}!x \quad p!y_2 > | < \boldsymbol{r^*} \quad \boldsymbol{q^*} \quad p^*!y_3 > | < p!y_3 \quad \boldsymbol{q!z_1} \quad \boldsymbol{r} \quad \boldsymbol{q^*!z_1} \quad p^*!y_2 \quad t^{\hat{}*}!x >$$
$$\xrightarrow{(R3)} < t\hat{}!x \quad p!y_2 > | < \boldsymbol{r^*!u} \quad \boldsymbol{q^*!z_2} \quad p^*!y_3 > | < p!y_3 \quad \boldsymbol{q!z_2} \quad \boldsymbol{r!u} \quad \boldsymbol{q^*} \quad p^*!y_2 \quad t^{\hat{}*}!x >$$

*3.1. Strand Graph*

In the previous section we have described the syntax and semantics of process calculus which is used to model, simulate and analyze the concurrent communicating processes of DNA computation. This formal language is widely appreciated for dealing with strand displacements in DNA sequences having secondary structures, like, branches, loops etc. But there are some limitations of process calculus in implementation of different rules, because of the complexity of pattern matching on arbitrary process contexts. To overcome this problem Petersen *et. al.* [Petersen *et. al.*, 2016] introduces the concept of *strand graph*.

Graphs are mathematical structures which are used to model pair-wise relations between objects. The graphical structures are formed by vertices or nodes which are connected by edges. In a graph if there is no distinction between the two nodes associated with each edge, the graph is said to be undirected. In directed graph each edge has a specific direction from one node to another. In strand graph the expressive power of graph theory can represent rich secondary structures of DNA strands and implement the complex rules. Now we will summarize the notation for strand graph theory as demonstrated in the paper [Petersen *et. al.*, 2016].

Strand graph is defined by *G = (V, length, colour, A, toehold, E)*, where,

$V = \{1,……, N\}$ denotes the *set of vertices* of the graph. Each vertex, shown by natural number, represents a *DNA strand*. There are different *sites* in a vertex. Each *site s* denotes a specific *domain* of that strand. The vertices are drawn as circular arrow with a specific direction i.e. from 5' to 3' of a DNA strand. The sites are placed in a vertex according to the occurrences of the corresponding domain in the specific strand. Site is represented as $s = (s, n)$, where $v$ is a vertex and $n$ is the position of site $s$ in vertex $v$. Both $v$ and $n$ are natural numbers.

*length:* denotes a *function* which assigns a *specific length to each vertex*. Lengths are represented by natural numbers.

*colour:* denotes a *function* which assigns a *specific colour to each vertex*. Colours are also represented by natural numbers. Thus, it would be easier to identify a particular vertex representing a specific DNA strand. Colour is actually a function of the length. If $v_1$ and $v_2$ are two vertices of a strand graph, then, $length(v_1) = length(v_2) \Rightarrow colour(v_1) = colour(v_2)$.

*A* is the set of *admissible edges* of the strand graph. If two domains of the DNA strands are complementary, they are able to hybridize with each other by forming a bond. Then an edge can be drawn between the sites of the vertices representing those domains. Throughout the performance of the whole program, all bonds those are allowed to be formed are represented by the set of admissible edges. Edge is represented as $e = \{s_1, s_2\}$ where $s_1$ and $s_2$ are two sites and $s_1 \neq s_2$. Again, we can write that, $e = \{(v_1, n_1), (v_2, n_2)\}$.

*Toehold* is a function that returns true if admissible edges exist between the short domains i.e. toehold domains and returns false for admissible edges between the long domains.

*E* is the set of current edges of the strand graph which is expressed as $\{e_1, ....., e_I\} \subseteq A$. In the contrary of other above mentioned information to define strand graph, *E* is non-static information. During the execution of the program the set of current edges changes with the change in reduction rules. A domain in a DNA strand cannot bind with more than one domain at any given instant i.e. only one edge can be drawn from a given site at that point of time. This is can be expressed as, $(i \neq j) \Rightarrow e_i \cap e_j = \emptyset$.

Now, we will illustrate the representation of DNA strand graph by an example. Fig. 7 shows the mechanism of toehold-mediated four-way strand displacement and branch migration. This mechanism consists of four DNA strands. In this program two partially double stranded DNA sequences simultaneously exchange the strands. Four-way strand displacement method is initiated by unhybridized toehold domains. The intermediate structure of this program is called Holliday junction.

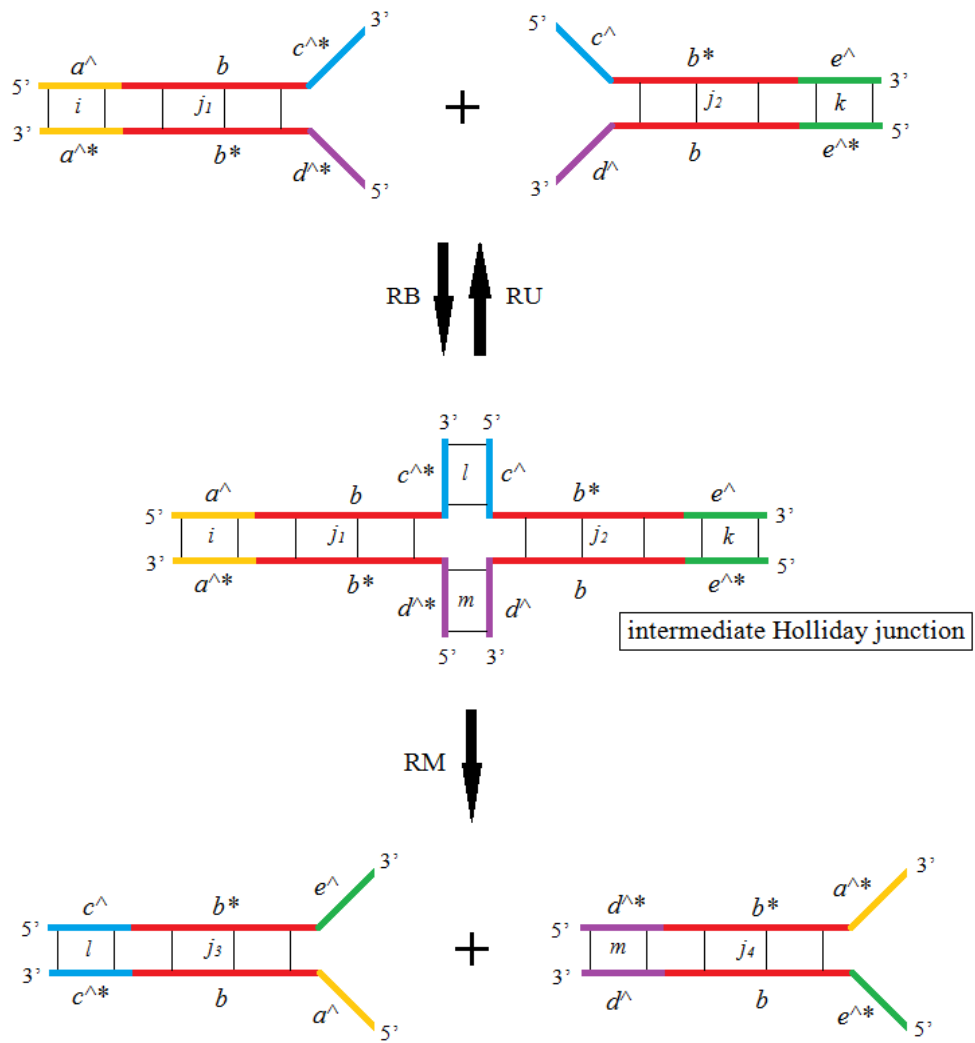

Figure 7. Mechanism of toehold-mediated four-way strand displacement and branch migration

The program codes of the DNA strands of the above described mechanism in the initial state are formed by process calculus. The codes of four strands are given below;

Strand 1 $<S_1>\ :\ <a^\wedge!i\ \ b!j_1\ \ c^{\wedge*}>$
Strand 2 $<S_2>\ :\ <d^{\wedge*}\ \ b^*!j_1\ \ a^{\wedge*}!i>$
Strand 3 $<S_3>\ :\ <c^\wedge\ \ b^*!j_2\ \ e^\wedge!k>$
Strand 4 $<S_4>\ :\ <e^{\wedge*}!k\ \ b!j_2\ \ d^\wedge>$

The DNA strand graph representing the initial state of toehold-mediated four-way strand displacement and branch migration mechanism is shown in Fig. 8.

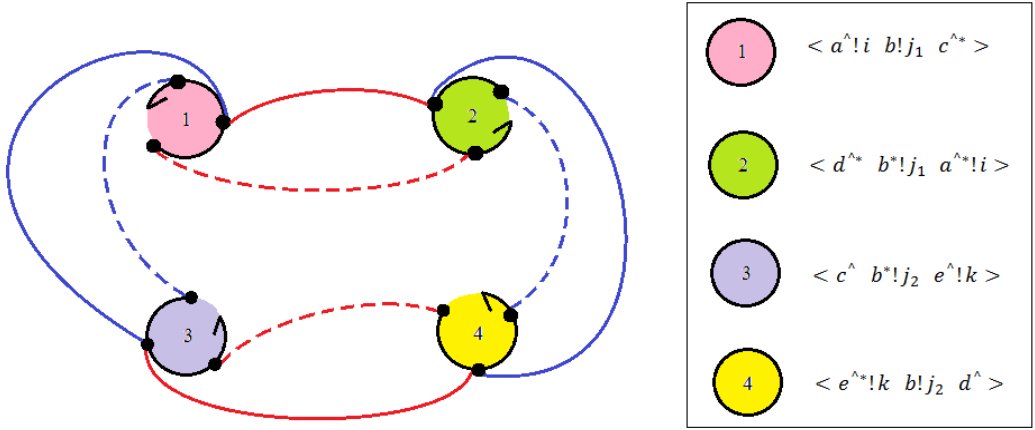

Figure 8. DNA strand graph *G* representing the initial state of toehold-mediated four-way strand displacement and branch migration

Each DNA strand in the program of toehold-mediated four-way strand displacement and branch migration (Fig. 7) is represented by vertex in the DNA strand graph as shown in Fig. 8. The arrowheads of the vertices which are drawn as circular arrows indicate the 3' end of the DNA strand. Different arbitrary colours are assigned for the vertices in the graph. For example, vertex 1 which represents the strand type $< a\hat{}\,!i \quad b!j_1 \quad c^{\hat{}\,*} >$ is assigned the colour pink. The domains of the DNA strands are presented by the sites which are placed on the vertices according to their occurrences. All the admissible edges are drawn in the strand graph. The current edges are represented by red lines and rests of the edges are represented by blue lines. The toehold edges are drawn by dashed lines.

The strand graph as shown in Fig. 8 is defined by *G = (V, length, colour, A, toehold, E)*, where,

| | | |
|---|---|---|
| *V* | = | {1, 2, 3, 4}. |
| *length* | = | {1 → 3, 2 → 3, 3 → 3, 4 → 3}. |
| *colour* | = | {1 → 1, 2 → 2, 3 → 3, 4 → 4}. |
| *A* | = | {(1, 1), (2, 3)}, {(1, 2), (2, 2)}, {(1, 2), (3, 2)}, {(1, 3), (3, 1)}, {(2, 1), (4, 3)}, {(2, 2), (4, 2)}, {(3, 2), (4, 2)}, {(3, 3), (4, 1)}. |
| *toehold* | = | {{(1, 1), (2, 3)} → *true*, {(1, 3), (3, 1)} → *true*, {(2, 1), (4, 3)} → *true*, {(3, 3), (4, 1)} → true, other → false}. |
| *E* | = | {(1, 1), (2, 3)}, {(1, 2), (2, 2)}, {(3, 2), (4, 2)}, {(3, 3), (4, 1)}. |

Now, we will illustrate some functions which are used to define DNA strand graph. Let, the program of toehold-mediated four-way strand displacement and branch migration is denoted by *P*.

$P = < S_1 > | < S_2 > | < S_3 > | < S_4 >$
$\quad = < a\hat{}\,!i \quad b!j_1 \quad c^{\hat{}\,*} > | < d^{\hat{}\,*} \quad b^*!j_1 \quad a^{\hat{}\,*}!i > | < c^{\hat{}} \quad b^*!j_2 \quad e^{\hat{}}!k > | < e^{\hat{}\,*}!k \quad b!j_2 \quad d^{\hat{}} >$

The DNA strand graph representing program *P* is *G = (V, length, colour, A, toehold, E)*. The function *tp* omits all the bonds from a specific domain. For example $tp(b!j_1) = b$.

The first strand, $<S_1> = <a^{\wedge}!i \ \ b!j_1 \ \ c^{\wedge*}>$ has three domains. It can be written that the type $tp(S_1) = tp(a^{\wedge}!i), tp(b!j_1), tp(c^{\wedge*}) = a^{\wedge}, b, c^{\wedge*}$. The function representing the length of $S_1$ is $len(S_1) = 3$.

The strand types are numbered according to their appearance in the given program (for example, $t_1, t_2, t_3, t_4$) depending on which the colour function is defined.

The domain function *dom* indicates a specific domain of the strand graph. For the DNA strand graph *G* corresponding to program *P*, *dom*(2, 3) indicates the 3$^{rd}$ domain of $S_2$. Another domain function *ndom*(2, 3) indicates the name of *dom*(2, 3) after omitting all bonds.

The toehold function *toe* defines the toehold of a specific DNA strand. For strand graph *G*, *toe*(3, 1) returns true which indicates *ndom*(3, 1) is a toehold domain.

*A* is the set of admissible edges of *G*. {(3, 3), (4, 1)} is the edge joining the 3$^{rd}$ domain of $S_3$ and the 1$^{st}$ domain of $S_4$ and {(3, 3), (4, 1)} ∈ *A*. Then, it can be written that, $(3, 3) \overset{A}{\leftrightarrow} (4, 1)$.

The following definition can be written to define a DNA strand graph [Petersen *et. al.*, 2016] using the above explained function;

| | | | |
|---|---|---|---|
| *V* | = | {1, ....., *N*} | where, *N* is natural number |
| *length(v)* | = | *len(Sv)* | |
| *colour(v) = i* | ⇔ | *tp(Sv) =$t_i$* | |
| $(v_1, n_1) \overset{A}{\leftrightarrow} (v_2, n_2)$ | ⇔ | $ndom(v_1, n_1) = comp(ndom(v_2, n_2))$ | |
| *toehold*({$s_1, s_2$}) | ⇔ | $toe(s_1)$ | |
| $(v_1, n_1) \overset{E}{\leftrightarrow} (v_2, n_2)$ | ⇔ | $\exists d, j. \ dom(v_1, n_1) = d!j \wedge dom(v_2, n_2) = comp(d)!j$ | |

where, *d* denotes the domain and *j* denotes the bond between $(v_1, n_1)$ and $(v_2, n_2)$.

In the next section we will illustrate the semantics of reduction rules.

*3.1.1. Semantics of reduction rules*

DNA strand graph transits from one state to another by following the reduction rules. The change in state of the strand graph is indicated by the change in colours of the edges among vertices. The semantics of the reduction rules need definitions of few functions [Petersen *et. al.*, 2016].

The function *sites(E)* returns the set of sites in set of current edges *E* which can be expressed by $\{s | \exists e \in E. s \in e\}$.

If two edges in a strand graph not only exist between the same pair of vertices but also the corresponding sites are adjacent to each other, the two edges are said to be adjacent. The function *adjacent(e, E)* returns the set of adjacent edges to edge *e* from the set *E*.

The function *hidden(e, E)* returns true if one of the ends of edge *e* from the set *E* occurs within a closed loop.

The function *anchored(e, E)* returns true if the edge *e* from the set *E* is a part of a stable junction by holding the corresponding sites close to each other.

Now we will describe the semantics of reduction rules through which the program occurs and reaches to its final state.

*Rule (GB)*

Let the sites of two vertices of a DNA strand graph is joined by admissible edge *x* which is not current at that instant. If those two sites are not preoccupied and open to each other, according to *rule (GB) x* can be converted into current edge. The semantics of rule (GB) is given below;

$$(GB) \frac{x \in A \backslash E \quad x \cap sites(E) = \emptyset \quad \neg hidden(x, E)}{E \xrightarrow{GB,\{x\}} E \cup \{x\}}$$

*Rule (GU)*

Let the sites of two vertices of a DNA strand graph is joined by admissible edge *e* and the sites represent toehold domain. Toehold domains are short enough to spontaneously unbind from its complement. Thus according to *rule (GU)* if the toehold domains are not anchored, the edge *e* can be removed from the current set *E* of the corresponding strand graph. The semantics of rule (GU) is given below;

$$(GU) \frac{e \in E \quad toehold(e) \quad \neg anchored(e, E)}{E \xrightarrow{GU,\{e\}} E \backslash \{e\}}$$

*Rule (G3)*

Let the sites of two vertices of a DNA strand graph is joined by admissible edge *x* which is not current at that instant. *x* can be joined to the set of current edges *E* even though one of the end sites is preoccupied by some other site forming a current edge *e*. *x* becomes current edge by removing *e* if the function *anchored(x, E)* returns true. This mechanism is termed as *displacing path*. The swapping of single bonds can form a long chain through the whole program. This mechanism is performed by reduction *rule (G3)*. The semantics of rule (G3) is given below;

$$(G3) \frac{e \in E \quad x \in A \backslash E \quad e = \{s, s'\} \quad x = \{s, s''\} \quad s'' \notin sites(E) \quad anchored(x, E)}{E \xrightarrow{G3,\{x\}} (E\{e\}) \cup \{x\}}$$

*Rule (GM)*

By the reduction *rule (GM)* the mechanism of displacing path i.e. swapping of single bonds makes a loop. The semantics of rule (GM) is given below;

$$(GM) \frac{i \in \{1,\dots,N\} \quad e_i \in E \quad x_i \in A \backslash E \quad e_i = \{s_i, s'_i\} \quad x_i = \{s'_{i-1}, s_i\} \quad s'_0 = s'_N \quad anchored(x_i, E)}{E \xrightarrow{GM,\{x_1,\dots,x_N\}} (E \backslash \{e_1,\dots,e_N\}) \cup \{x_1,\dots,x_N\}}$$

*3.1.2. Graphical illustration of reduction rules*

In Fig. 7 the entire mechanism of toehold-mediated four-way strand displacement and branch migration, which is graphically interpreted in Fig. 8, is shown. In this section we will pictorially describe (Fig. 9) how the reduction rules work in DNA strand graph *G* as shown in Fig. 8.

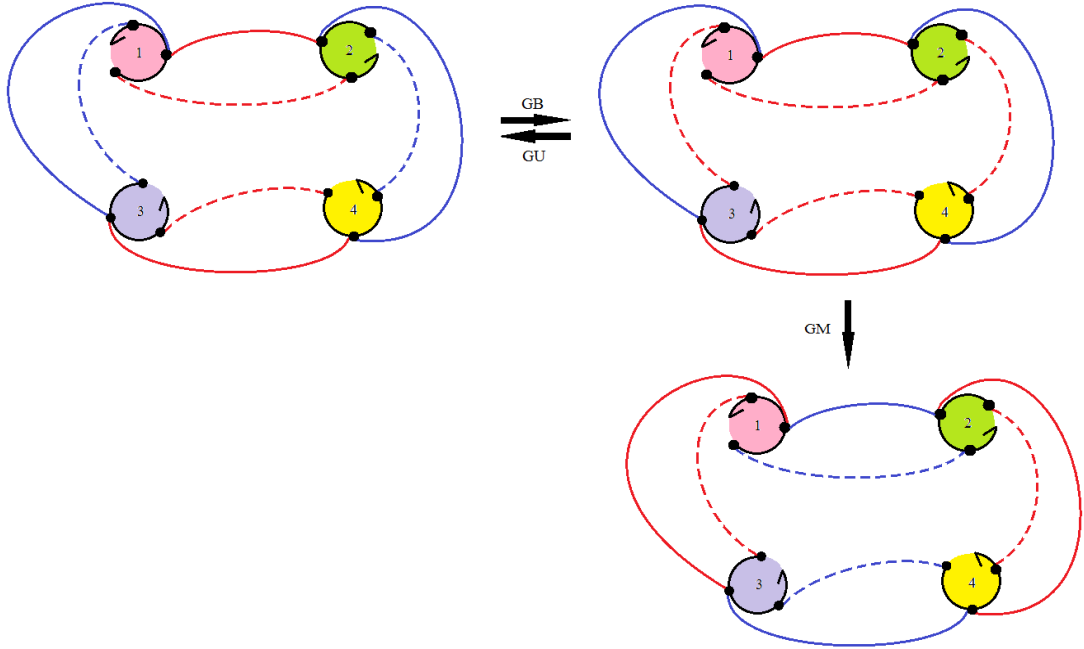

Figure 9. DNA strand graph with reduction rules conducting the program of toehold-mediated four-way strand displacement and branch migration

In section 2 of this paper we have discussed theorem proving with resolution refutation in propositional logic. Section 3 illustrates the syntax and semantic of process calculus and DNA strand graph. In next section we will show how theorem proving with resolution refutation can be performed by DNA computation. We will implement process calculus and strand graph in the domain of theorem proving.

## 4. Theorem Proving Based on Process Calculus and DNA Strand Graph Semantics

Resolution refutation or proof by contradiction proves a theorem by negating the statement to be proved. If the theorem prover, generated by propositional resolution, proves the inconsistency of the negated goal with the given set of premises, this implies that the original goal is consistent. Thus, the principle of resolution refutation is used to prove the unsatisfiability of a set of clauses. In Fig. 1 of section 2 we have shown the deduction tree of a set *S* containing six clauses.

$$\left.\begin{array}{l}(i)\ P \vee \neg Q \vee R \\ (ii)\neg U \vee V \vee \neg R \\ (iii)\ Q \\ (iv)\neg V \\ (v)\neg P \\ (vi)\ U\end{array}\right\} S$$

In the subsection 4.1 we will encode the above mentioned set of clauses in terms of DNA strands. We will show how theorem proving by resolution refutation can be performed in DNA computation using DNA strands and elementary operations to manipulate the strands. In subsection 4.2 the theorem proving is coded by process calculus. Subsection 4.3 shows the representation of the entire program by DNA strand graph and reduction rules.

### *4.1. Theorem proving by resolution refutation in DNA computation*

Lee, Park, Jang, Chai and Zhang [*Lee et. al.*, 2002] performed theorem proving by resolution refutation using DNA strands with the help of some elementary operations to manipulate the strands. To prove the unsatisfiability of the set of clauses $S$ by resolution refutation in DNA computation, few steps should be followed.

### *Step 1.*

The clauses of set $S$ contain five propositional variables or literals. Each literal is encoded by arbitrarily chosen ten bases long single-stranded DNA oligonucleotide. The negation of each literal is encoded by the complementary sequence of the corresponding DNA strand. The encoded single-stranded DNA oligonucleotides are listed in Table 1.

| Literal | Encoded DNA strand |
|---|---|
| $P$ | $5' -$ ACGTAGTCAC $- 3'$ |
| $\neg P$ | $3' -$ TGCATCAGTG $- 5'$ |
| $Q$ | $5' -$ CAGTCAATTC $- 3'$ |
| $\neg Q$ | $3' -$ GTCAGTTAAG $- 5'$ |
| $R$ | $5' -$ TCAGTCGAAT $- 3'$ |
| $\neg R$ | $3' -$ AGTCAGCTTA $- 5'$ |
| $U$ | $5' -$ CTAGGTCCAT $- 3'$ |
| $\neg U$ | $3' -$ GATCCAGGTA $- 5'$ |
| $V$ | $5' -$ GATCGTGCAT $- 3'$ |
| $\neg V$ | $3' -$ CTAGCACGTA $- 5'$ |

Table 1. Representation of literals by DNA strands

### *Step 2.*

All the clauses are encoded in terms single-stranded DNA oligonucleotides. To encode the clauses, the DNA strands representing literals of the corresponding clause are concatenated. The encoded clauses are listed in Table 2.

| Clause | Encoded DNA strand |
|---|---|
| $P \lor \neg Q \lor R$ | $5' - \overbrace{\text{ACGTAGTCAC}}^{P} \overbrace{\text{GAATTGACTG}}^{\neg Q} \overbrace{\text{TCAGTCGAAT}}^{R} - 3'$ |
| $\neg U \lor V \lor \neg R$ | $5' - \overbrace{\text{ATGGACCTAG}}^{\neg U} \overbrace{\text{GATCGTGCAT}}^{V} \overbrace{\text{ATTCGACTGA}}^{\neg R} - 3'$ |
| $Q$ | $5' - \text{CAGTCAATTC} - 3'$ |
| $\neg V$ | $5' - \text{ATGCACGATC} - 3'$ |
| $\neg P$ | $5' - \text{GTGACTACGT} - 3'$ |
| $U$ | $5' - \text{CTAGGTCCAT} - 3'$ |

Table 2. Representation of clauses by DNA strands

*Step 3.*

All the single-stranded DNA oligonucleotides as shown in Table 2 are mixed in a test tube and allowed to hybridize with each other. In this step the principle of resolution refutation is implemented by DNA strand hybridization. The DNA strands encoding the clauses hybridize with each other to generate resolvent. The resolvent may be partially double-stranded or full double-stranded DNA sequence. Fully double-stranded DNA sequence denotes empty clause { }.

*Step 4.*

The resultant hybridized DNA strands are allowed to be ligated using specific ligase enzyme.

*Step 5.*

The ligated sequences obtained from step 4 are amplified using specific primers by polymerase chain reaction. The primers are chosen specifically so that unwanted sequences are not amplified.

*Step 6.*

Gel electrophoresis is performed to verify whether fully double-stranded DNA sequence is present in the resultant amplified sequences. Fully double-stranded DNA sequence denotes empty clause { }.

If an empty clause { } is derived from the set of clauses *S* by resolution, it is proved that *S* is unsatisfiable. Thus, the fully double-stranded resultant DNA sequence (resolvent) establishes the unsatisfiability of *S*. The given theorem is proved by contradiction. If all the resultant sequences are single-stranded or partially double-stranded, then it is proved that *S* is satisfiable.

Fig. 10 is the pictorial representation of the process of theorem proving by resolution refutation with DNA strands. We also compare the process with deduction tree shown in Fig. 1 of section 2.

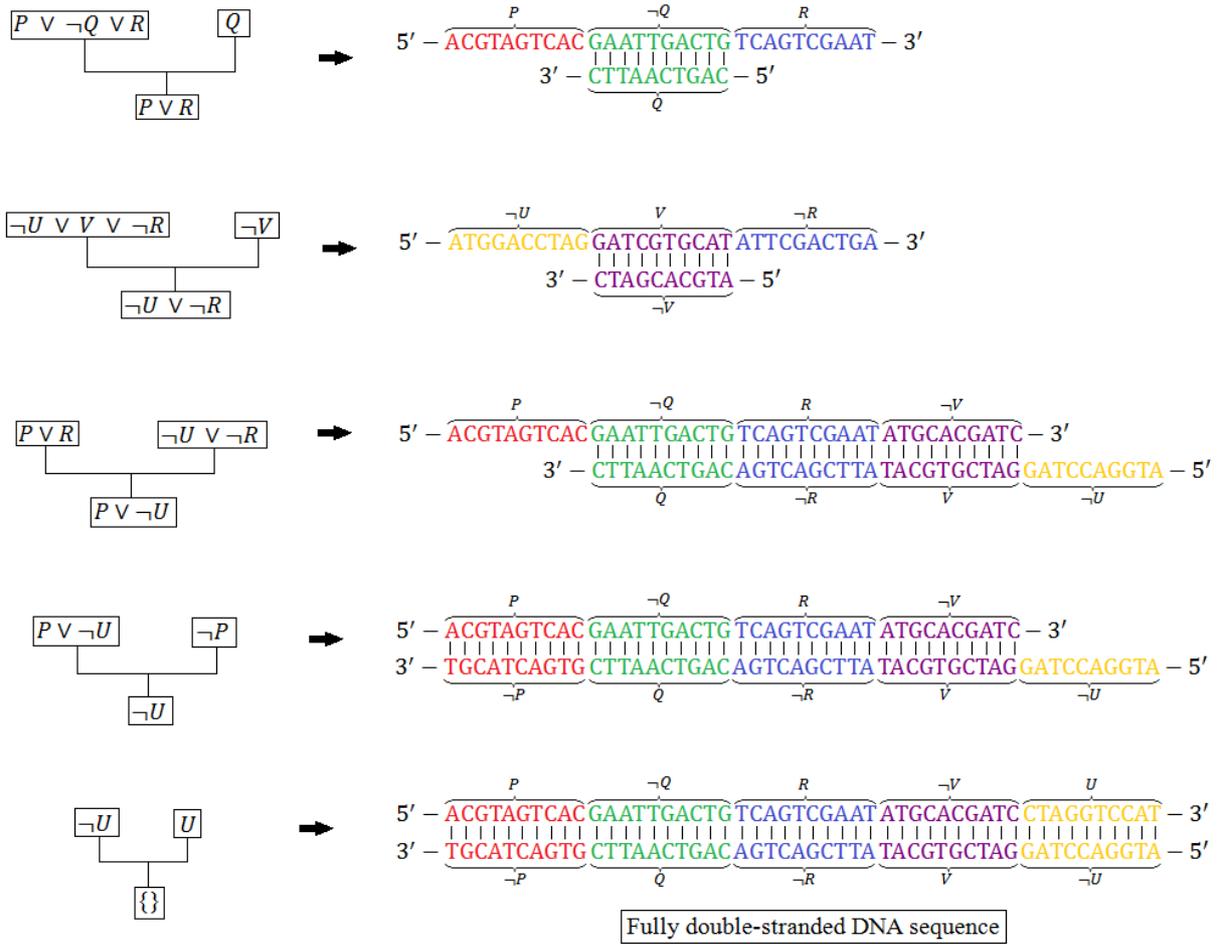

Figure 10. Representation of the process of theorem proving by resolution refutation with DNA strands and comparison with deduction tree

*4.2. Theorem proving by resolution refutation coded by process calculus*

In this section we will code theorem proving by resolution refutation by process calculus using the syntax and semantics described in section 3. We have to prove the unsatisfiability of the set of clauses *S*.

Let the entire program is denoted by *P*. The program *P* consists of six clauses which are encoded by single-stranded DNA oligonucleotides. *P* is defined as the multiset of six DNA strands.

$$P = <S_1> | <S_2> | <S_3> | <S_4> | <S_5> | <S_6>$$

Therefore,

$$P = < P \quad Q^* \quad R > | < U^* \quad V \quad R^* > | < Q > | < V^* > | < P^* > | < U >$$

Every literal is encoded by arbitrarily chosen ten bases long single-stranded DNA sequence. The DNA strand encoding the negation of each literal is the Watson-Crick complement of the corresponding literal. Thus, the DNA strand encoding the negation of literal $P$, i.e. ¬$P$, is named as $P^*$. For all the literals the same rule has been followed. Each of the strands <$S_1$> and <$S_2$> contains three domains as given by the program code. The remaining strands, i.e. <$S_3$>, <$S_4$>, <$S_5$> and <$S_6$>, contain one domain each. From the program code it is clear that at the initial state of the program all the domains of the strands are free. As the domain $Q^*$ of <$S_1$> and the domain $Q$ of <$S_3$> are not bound with any other domain, the program matches the context $C(Q, Q^*)$. It can be written that $P=C(Q!i, Q^*!i)$ as one end of the bond $i$ is not in closed loop, i.e. *hidden*($i$, $P$) returns false. Thus, the program $P'$ can be produced by the *rule* (*RB*) which forms the new bond $i$ between the second domain of <$S_1$> and the only domain of <$S_3$>. The program code is given below;

$$< P \quad Q^* \quad R > \;|\; < U^* \quad V \quad R^* > \;|\; < Q > \;|\; < V^* > \;|\; < P^* > \;|\; < U >$$

$$\xrightarrow{(RB)} <P \quad Q^*!i \quad R> \;|\; <U^* \quad V \quad R^*> \;|\; <Q!i> \;|\; <V^*> \;|\; <P^*> \;|\; <U>$$

As the domains $V$ of <$S_2$> and $V^*$ of <$S_4$> are free, the program matches the context $C(V, V^*)$. The new bond $j$ can be formed joining these two domains by the *rule* (*RB*) as one end of the bond is not hidden. The program code is given below;

$$< P \quad Q^*!i \quad R > \;|\; < U^* \quad V \quad R^* > \;|\; < Q!i > \;|\; < V^* > \;|\; < P^* > \;|\; < U >$$
$$\xrightarrow{(RB)} <P \quad Q^*!i \quad R> \;|\; <U^* \quad V!j \quad R^*> \;|\; <Q!i> \;|\; <V^*!j> \;|\; <P^*> \;|\; <U>$$

Now, the second domains of the strands <$S_1$> and <$S_2$> are bound. The domains $R$ and $R^*$ at 3' ends of the strands <$S_1$> and <$S_2$> are free. Thus, the program matches the context $C(R, R^*)$. The new bond $k$ can be formed between these domains by *rule* (*RB*). The ends of the bond do not occur in closed loop. The program code is given below;

$$< P \quad Q^*!i \quad R > \;|\; < U^* \quad V!j \quad R^* > \;|\; < Q!i > \;|\; < V^*!j > \;|\; < P^* > \;|\; < U >$$
$$\xrightarrow{(RB)} <P \quad Q^*!i \quad R!k> \;|\; <U^* \quad V!j \quad R^*!k> \;|\; <Q!i> \;|\; <V^*!j> \;|\; <P^*> \;|\; <U>$$

Again, *rule* (*RB*) comes into action and a new bond $l$ is formed between the free domains $P$ of <$S_1$> and $P^*$ of <$S_5$> as one end of the $l$ is not hidden. The program code is given below;

$$< P \quad Q^*!i \quad R!k > \;|\; < U^* \quad V!j \quad R^*!k > \;|\; < Q!i > \;|\; < V^*!j > \;|\; < P^* > \;|\; < U >$$
$$\xrightarrow{(RB)} <P!l \quad Q^*!i \quad R!k> \;|\; <U^* \quad V!j \quad R^*!k> \;|\; <Q!i> \;|\; <V^*!j> \;|\; <P^*!l> \;|\; <U>$$

Except the domains $U^*$ of <$S_2$> and $U$ of <$S_6$>, all the domains of program $P$ is bound. The program matches the context $C(U, U^*)$. New bond $m$ can be formed between these two domains by *rule* (*RB*). The program code is given below;

$$< P!l \quad Q^*!i \quad R!k > \;|\; < U^* \quad V!j \quad R^*!k > \;|\; < Q!i > \;|\; < V^*!j > \;|\; < P^*!l > \;|\; < U >$$
$$\xrightarrow{(RB)} <P!l \quad Q^*!i \quad R!k> \;|\; <U^*!m \quad V!j \quad R^*!k> \;|\; <Q!i> \;|\; <V^*!j> \;|\; <P^*!l> \;|\; <U!m>$$

Form the program code it is clear that all the domains of the given program are bound. Thus, the resultant strand is complete double stranded DNA sequence which indicates empty clause {}. This proves the unsatisfiability of the set of clauses *S*.

### *4.3. Theorem proving by resolution refutation using DNA strand graph and reduction rules*

This section is the graphical representation of program *P* which has been described by process calculus using program codes in the previous section. The unsatisfiablity of the set of clauses *S* is demonstrated using DNA strand graph *T*. Initially the code of program P is represented by the expression,

$$P = < P \quad Q^* \quad R > | < U^* \quad V \quad R^* > | < Q > | < V^* > | < P^* > | < U >$$

Graphical depiction of program *P* is shown in Fig. 11.

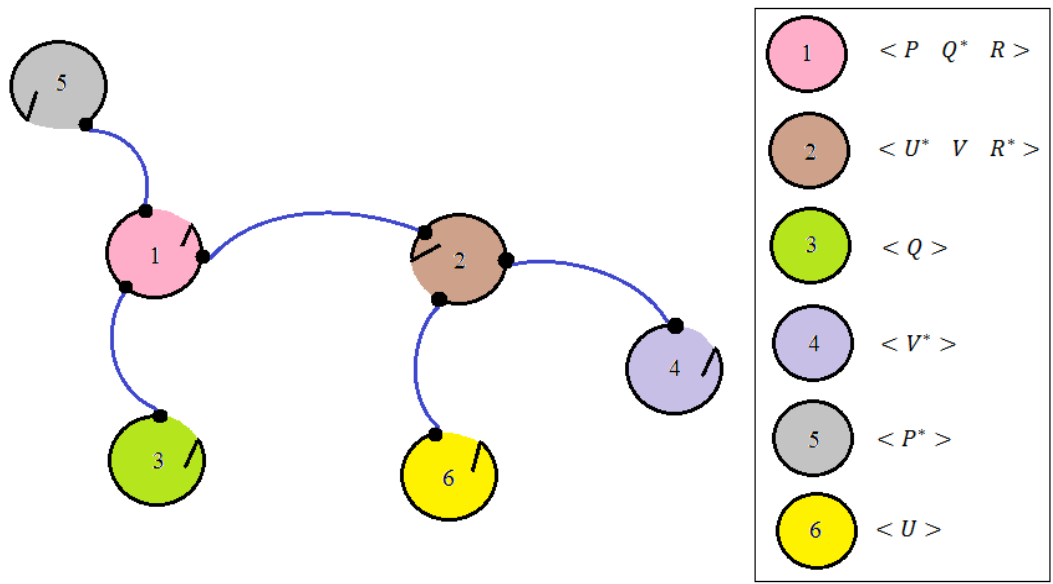

Figure 11. DNA strand graph *T* representing the initial state of program *P*

Six strands of *P* is represented by six vertices in *T* (Fig. 11). Different arbitrary colours are assigned for the vertices in the graph. The domains of the DNA strands are presented by the sites which are placed on the arrow-headed vertices according to their occurrences. All the edges of the strand graph *T* are admissible edges. Since, at the starting point of the program all the DNA sequences are single stranded i.e. initially the set of current edges is empty i.e. $E = \emptyset$. The admissible edges are drawn by blue lines.

The initial state of DNA strand graph as shown in Fig. 11 is defined by *T = (V, length, colour, A, toehold, E)*, where,

| | | |
|---|---|---|
| *V* | = | {1, 2, 3, 4, 5, 6}. |
| *length* | = | {1 → 3, 2 → 3, 3→ 1, 4 → 1, 5 → 1, 6 → 1}. |
| *colour* | = | {1 → 1, 2 → 2, 3→ 3, 4 → 4, 5 → 5, 5 → 6}. |
| *A* | = | {(1, 1), (5, 1)},{(1, 2), (3, 1)},{(1, 3), (2, 3)},{(2, 1), (6, 1)},{(2, 2), (4, 1)}. |
| *toehold* | = | ∅. |
| *E* | = | ∅. |

In Fig. 12 the entire mechanism of theorem proving by resolution refutation using DNA strands is represented by strand graph *T* and reduction rules.

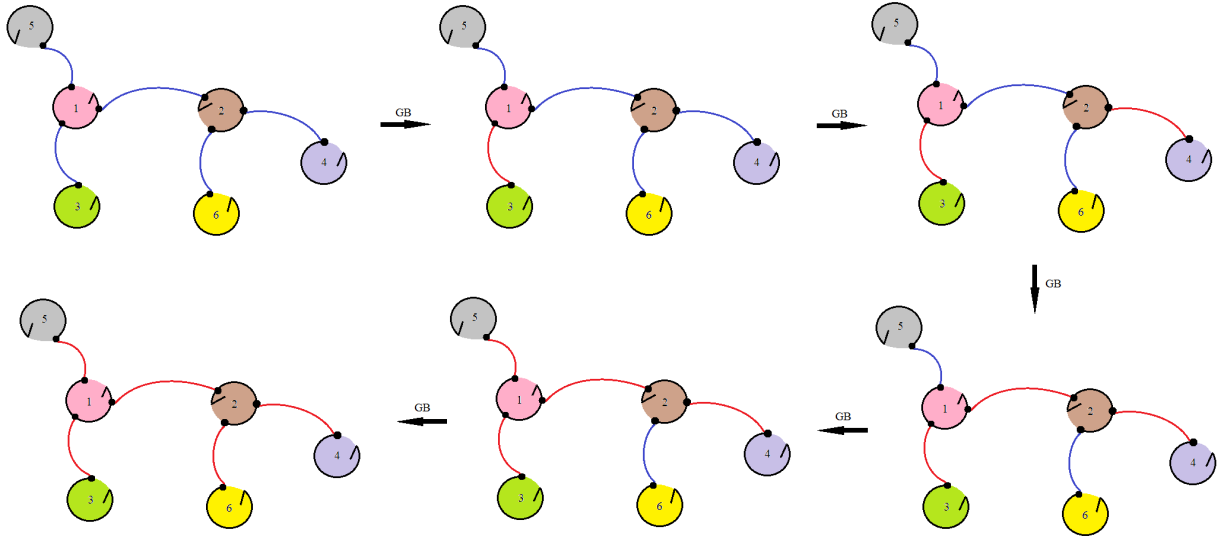

Figure 12. DNA strand graph with reduction rules conducting the program of theorem proving by resolution refutation using DNA strands

In Fig. 12, the admissible edges are drawn by blue lines. Initially the set of current edges *E* is empty, thus, all the edges are admissible. The $2^{nd}$ site (domain $Q^*$) of vertex 1 and the only site (*Q*) of vertex 3 are not preoccupied and open to each other. Thus, according to *rule (GB)* in the first step of the program the admissible edge joining these two sites is converted into current edge. The current edge is drawn by red lines. All the remaining admissible edges of the strand graph are converted into current edges in next few steps following the reduction rule (GB). Finally, all the edges of the graph are included in set *E*. No sites in graph is free in the resultant graphical structure. This indicates that, the final strand is complete double stranded DNA sequence which implies empty clause {}. Thus, the unsatisfiability of the set of clauses *S* has been proved.

## 5. Conclusion

In this paper we have shown how the theorem proving with resolution refutation by DNA computation can be presented by the semantics of DNA strand graph. The chemical potential and flexibility of DNA strands have been exploited to model theorem proving by resolution refutation. Formal language theory in form of process calculus has been successfully used as a tool for modeling and analyzing DNA operations performed for theorem proving in propositional logic. The model based on formal language theory is efficiently interpreted by DNA strand graph for better simulation. This new approach can be further extended for designing and modeling other expert systems based on first-order logic and fuzzy propositional logic.